\documentclass[10pt,conference,a4paper]{IEEEtran}
\usepackage{cite}
\usepackage{graphicx}
\DeclareGraphicsExtensions{.eps,.png,.jpg,.pdf}
\ifCLASSINFOpdf
\else
\fi
\usepackage{amsmath}
\usepackage{mathtools}
\usepackage{amsfonts}

\usepackage{algorithmic}
\usepackage{array}
\usepackage{float}
\ifCLASSOPTIONcompsoc
  \usepackage[caption=false,font=normalsize,labelfont=sf,textfont=sf]{subfig}
\else
  \usepackage[caption=false,font=footnotesize]{subfig}
\fi
\usepackage{url}
\begin{document}
\title{Spectral Image Visualization Using Generative Adversarial Networks}

\author{
\IEEEauthorblockN{Siyu Chen, Danping Liao, Yuntao Qian}
\IEEEauthorblockA{College of Computer Science\\Zhejiang University, HangZhou, China\\
sychen@zju.edu.cn, liaodanping@gmail.com, ytqian@zju.edu.cn}
%\and
%\IEEEauthorblockN{Liao Danping}
%\IEEEauthorblockA{College of Computer Science\\Zhejiang University, HangZhou, China\\
%liaodanping@gmail.com}
%\and
%\IEEEauthorblockN{Qian Yuntao}
%\IEEEauthorblockA{College of Computer Science\\Zhejiang University, HangZhou, China\\
%ytqian@zju.edu.cn}

}


%Telephone: (800) 555--1212, Fax: (888) 555--1212}


\maketitle

\begin{abstract}
Spectral images captured by satellites and radio-telescopes are analyzed to obtain information about geological compositions distributions, distant asters as well as undersea terrain. Spectral images usually contain tens to hundreds of continuous narrow spectral bands and are widely used in various fields. But the vast majority of those image signals are beyond the visible range, which calls for special visualization technique.
The visualizations of spectral images shall convey as much information as possible from the original signal and facilitate image interpretation.
However, most of the existing visualizatio methods display spectral images in false colors, which contradict with human's experience and expectation.
In this paper, we present a novel visualization generative adversarial network (GAN) to display spectral images in natural colors.
%Based on CycleGAN model and WGAN, we develop an end-to-end architecture using pixel-wise convolutions and residual learning technique.
%Unlike other methods that generate unnatural visualization which contradict with human's experience and expectation, or those produce natural visualizations by require pixel level matching for pixel referential color information, our approach is able to generate comprehensible visualizations in natural colors without the requirement of paired reference images or pixel-wise matching.
To achieve our goal, we propose a loss function which consists of an adversarial loss and a structure loss.
The adversarial loss pushes our solution to the natural image distribution using a discriminator network
that is trained to differentiate between false-color images and natural-color images.
We also use a cycle loss as the structure constraint to guarantee structure consistency.
Experimental results show that our method is able to generate structure-preserved and natural-looking visualizations.
\end{abstract}


% \fi
\IEEEpeerreviewmaketitle

\section{Introduction}

The rapid development of imaging technology brings about the ability to capture images with high spectral resolution.
Hyperspectral imaging sensors, for example, routinely capture more than $100$ channels of spectral data, while medical imaging systems capture multi-dimensional, and multi-modal image set.
Ultimately these images are often interpreted by human observers for analysis or diagnosis.
However, human's eyes are merely capable of sensing a very narrow range of the electromagnetic wavelength.
It is thus crucial to reduce the dimensionality of spectral images so that the images can be displayed on an output device.
Although the requirements of visualization are task dependent, there are some common goals such as information preservation, consistent rendering and natural palette~\cite{jacobson2005design}.

The most straightforward visualization method is to select three of the original bands to display.
Band selection methods such as
linear prediction (LP) was applied to select the most informative bands~\cite{su2014LP}.
One disadvantage of band selection methods is that, except for the selected bands, all the information contained in other channels is ignored.

To preserve the information across the whole range of wavelength, dimension reduction methods such as principle component analysis~\cite{tyo2003principal} and independent component analysis~\cite{zhu2007evaluation} were proposed.
To better preserve the nonlinear and local structures, manifold learning methods were applied to spectral image visualization tasks~\cite{bachmann2005exploiting}.%,crawford2011nonlinear}.
In \cite{kotwal2010visualization}, in order to preserve the edge information, bilateral filtering is applied to calculate the band weights at each pixel for band image fusion.
These local structure-based approaches demonstrate excellent performance in preserving the intrinsic information of spectral images.

While most existing visualization methods try to preserve as much information as possible from the original data,
most of them display spectral images in false colors which are hard to interpret when the colors of objects are very different from what is expected by humans.
Moreover, most data-adaptive methods suffer from ``inconsistent rendering'' problem, i.e., very different colors might be assigned to the same objects/materials in different images, which also hinders the interpretation of spectral images.
Therefore, ``natural palette'' and ``consistent rendering'' gradually become two important criteria for visualization quality evaluation.

To produce consistent natural-looking images, Jacobson \emph{et al}.~\cite{jacobson2005design} %,jacobson2007linear}
proposed a stretched color matching function (CMF) for visualization, which stretches the
CIE 1964 tristimulus color matching functions from the visible range to the invisible range.
However, the strecthed CMF is fixed for each type of hyperspectral imaging sensor, which limits its capability in preserving the specific information in differrent spectral images.

Connah \emph{et al}. ~\cite{Connah2014} used a constrained contrast mapping paradigm in the gradient domain to generate a visualization that shares similar colors with a corresponding natural-looking RGB image. This method requires pixel-wise matching between the spectrtal image and the corresponding RGB image, which limits its applications in general scenarios where pixel-wise matching is hard to obtain.

Liao \emph{et al}. utilized manifold alignment to transfer colors from natural RGB images to hyperspectral images~\cite{liao2016hscolor,liao2014hscolor,liao2013hscolor}.
The approach is capable of visualizing the hyperspectral image in natural colors as well as preserving local similarity between hyperspectral pixels.
However, manifold alignment also requires a set of matching pixels between the hyperspectral image and the referencing RGB image. As a result, the corresponding RGB image should be captured at exactly the same site as the hyperspectral image and should not have large geometric distortion in order to obtain precise image registration.

Recently, Generative Adversarial Network (GAN) based techniques have demonstrated to be very effective in colorization and style transfer~\cite{cyclegan,pix2pix,dcgan}. Unlike traditional style transfer methods that specifically and explicitly design objectives for the model, GAN utilizes a discriminator network to guide the generator. Such an adversarial learning technique makes it possible to train a strong generative model in an unsupervised way without pixel labeling or paired samples. These advantages make GAN a promising visualization technique.
\begin{figure}[!t]
\centering
\includegraphics[width=3.6in]{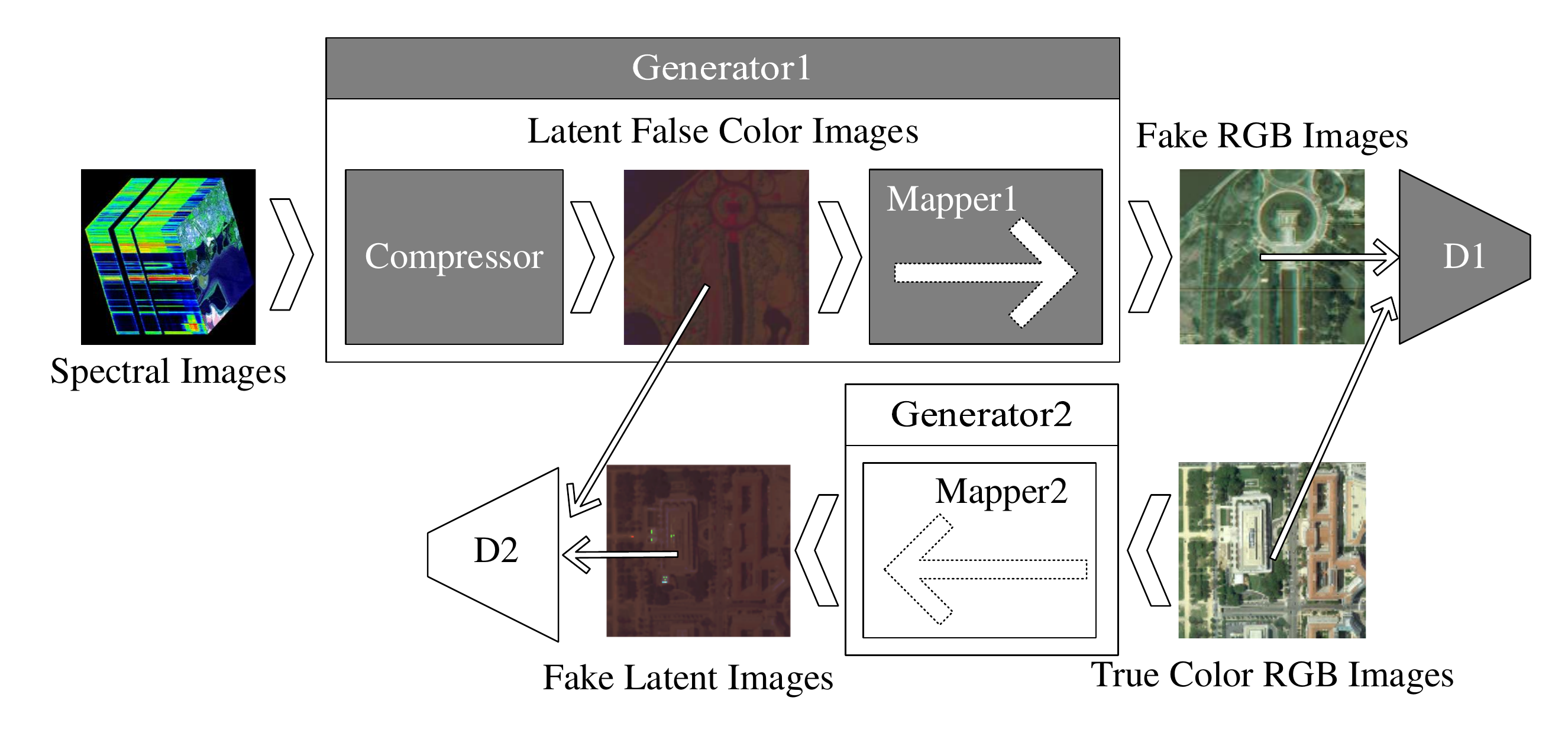}
\caption{Model structure.}
\label{fig:model}
\end{figure}

In this work we propose an end-to-end visualization generative adversarial network (VGAN), in which a deep residual network (ResNet) is used.
Our goal is to generate natural-looking and structure-preserved visualizations for spectral images.
Different from previous methods that require image-pairing and pixel-wise matching between natural images and the spectral images to ``transfer''  colors, our model is totally unsupervised and can automatically learn the correspondence between different data distributions.
To achieve our goal, we propose a loss function which consists of an adversarial loss and a cycle-consistency loss.
The adversarial loss pushes our solution to the natural image manifold using a discriminator network
that is trained to distinguish false-color images from natural-color images.
The cycle-consistency loss guarantees the structure of the spectral image to be preserved during color mapping.

%In VGAN the aim is to generate a natural-looking visualization $I$ for an input remote sensing spectral image $S$. In training, $I$ are satellite images obtained from Google Earth.
The model structure of the VGAN is shown in Fig. \ref{fig:model}.
Based on Cycle-GAN~\cite{cyclegan}, VGAN comprises two generators and two discriminators.
Generator1 contains: 1. a \textit{compressor} network that fuses input spectral images to a 3-band image; 2. a \textit{mapper} that translates the output of the compressor to a natural-looing image.
Discriminator1 encourages Generator1 to produce visualizations indistinguishable from natural color images.
We also use a second generator, a mapper from natural-looking images to the latent output space of the compressor, to guarantee structure consistency by minimizing the cycle loss.
%Generator2 is trained with the goal of fooling Discriminator2 that is trained to distinguish
%the output of Generator2 from the output of the compressor.
%Experiments show that the constraint imposed by the Generator2 is essential for generating structure-consistent visualizations.
Both compressor and mapper networks use very small convolution kernels which allows the feature maps to be more structural consistent. Also, using small size kernels makes the model faster and easier to train, as pointed out by recent model compression researches~\cite{squeezenet2016,mobilenet2017}.

This paper is organized in the following order: Section \ref{sec:gan} explains GAN and its adversarial learning techniques; Section \ref{sec:proposed_method} introduces the proposed GAN for spectral image visualization; Section \ref{sec:experiments} shows the experimental results and comparisons. The conclusions are presented in Section \ref{sec:conclusion}.

\section{Generative Adversarial Nets}
\label{sec:gan}

GAN was firstly introduced by Goodfellow \cite{goodfellow2014generative}, in which, noise $z$ sampled from uniform distribution is passed to an up-sampling network $G$ to generate image $G(z)$ from some latent distribution. The goal of network $G$ is to generate images that are highly similar to real images and thus difficult to classify by a differentiable discriminator $D$. This is achieved by maximizing the targets $L_D$ and $L_G$ alternately.
\begin{eqnarray}
  \label{eq:ld}
  L_D &=& \underset{\substack{ x\sim p_{data} \\ z\sim p_{noise}} }{\mathbb{E}}  \log(D(x)) + \log \left(1-D(G(z)) \right),\\
  \label{eq:lg}
  L_G &=&  \underset{z\sim p_{noise} }{\mathbb{E}} \log(D(G(z))) .
\end{eqnarray}

$D$ takes in real image $x\sim P_{data}$ and fake image $G(z)$ and gives discriminative confidence $y\in [0,1]$. Higher confidence indicates more likely of being a real image.

This adversarial process has an optimal solution where nash equilibrium is reached. The following equations hold when  the optimal solution is reached.
\begin{eqnarray}
  \underset{\substack{ x\sim p_{data} \\ z\sim p_{noise}} }{\mathbb{E}} \log(D^*(G^*(z))) - \log(D^*(x)) &=& 0 ,\\
  \underset{z\sim p_{noise} }{\mathbb{E}} D^*(G^*(z)) &=& 0.5 ,\\
  \underset{x\sim p_{data} }{\mathbb{E}} D^*(x) &=& 0.5 .
\end{eqnarray}

%The equilibrium means the generator generates delicate fakes which are indistinguishable from real images, thus yielding $50\%$ probability for both.

\section{Proposed method}
\label{sec:proposed_method}
The goal of VGAN is to estimate a structure-preserved and natural-looking visualization for an input spectral image.
In our model, a spectral image from domain $A$ is firstly transformed to domain $B$ by a compressor, which compresses the number of channels to 3 and outputs a false-color image.
Then the false-color image is mapped to domain $C$ where images have natural color distributions.
In order to generate structure-consistent visualizations, we impose a cycle loss  by using a second generator that maps images from domain $C$ to domain $B$ when training the model.

VGAN is constructed with heterogeneous network architecture for generators $G_{A\rightarrow C}$ and $G_{C\rightarrow B}$.
The generator $G_{A\rightarrow C}$ contains two parts: compressor $Cpr$ and mapper $M_1$. The generator $G_{C\rightarrow B}$ contains only one mapper $M_2$. The $M_1$, $M_2$ and $Cpr$ correspond with \textit{Mapper 1}, \textit{Mapper 2} and \textit{Compressor} in Fig. \ref{fig:model}.

\subsection{Formulation}
The objective of VGAN is to learn a mapping function from spectral image domain $A$ to natural RGB domain $C$ via a latent false-color domain $B$ as a bridge, given training samples ${\{x_i\}}_{i=1}^N$ where $x_i \in A$, and ${\{y_j\}}_{i=1}^M$ where $y_i \in C$. There are three mapping functions to be learned. One of the mapping functions maps the image from domain $A$ to $B$, one from $B$ to $C$ and another one maps the image from $C$ back to $B$. We denote the data distribution as $x\sim \mathbb{P}_A$, $y\sim \mathbb{P}_C$ and $z\sim \mathbb{P}_B$ where $z$ is translated from domain $A$. Formally, the three mappings are $Cpr:A\rightarrow B$, $M_1:B\rightarrow C$ and $M_2:C\rightarrow B$, which compose two generative mappings $G_{A\rightarrow C}(x)=M_1(Cpr(x))$ and $G_{C\rightarrow B}(y)=M_2(y)$. In addition, we use two adversarial discriminators $D_C$ and $D_B$, where $D_C$ aims to distinguish between images $\{y\}$ and visualized images $G_{A\rightarrow C}(x)$; in the same way, $D_B$ aims to discriminate between $\{z\}$ and $G_{C\rightarrow B}(y)$. Our objective contains two types of terms: \textit{adversarial losses} to constrain the output of mapping functions to be as similar as possible to the target domain; and \textit{cycle consistency losses} to preserve the structure.

\subsubsection{Adversarial Loss}
Adversarial losses are applied to both of the generators. For the generator $G_{A\rightarrow C}$ its discriminator $D_C$, the objective is formulated as follows:
\begin{multline}\label{eq:adv_loss}
  \mathcal{L}_{GAN}(G_{A\rightarrow C},D_C,A,C) = \underset{y\sim \mathbb{P}_C}{\mathbb{E}} \log(D_C(y)) \\
  + \underset{x\sim \mathbb{P}_A}{\mathbb{E}} \log(1-D_C(G_{A\rightarrow C}(x))),
\end{multline}
where $G_{A\rightarrow C}$ takes in images from domain $A$ and generate images similar to images from domain $C$, while $D_C$ aims to distinguish fake images generated by $G_{A\rightarrow C}$ from images in domain $C$. A similar adversarial loss is introduced for the generator $G_{C\rightarrow B}$ and its discriminator $D_B$ as well: i.e. $min_{G_{C\rightarrow B}}max_{D_B}\mathcal{L}_{GAN}(G_{C\rightarrow B},D_B,C,B)$ where the samples of domain $B$ is generated by $Cpr(x), x\sim \mathbb{P}_A$.

\subsubsection{Cycle Consistency Loss}
Because of the implications of adversarial loss that, given a mapping function $M$, any mappings obtained by permutating the images in the target domain gives similar loss value. Thus, with only adversarial loss, it cannot be guaranteed that the learned function can map input $x_i$ to a desired $y_i$.
Also, cycle consistency loss can help to prevent the function from mapping all the $x_i$ to a single highly realistic $y_i$, as have been reported in \cite{traingan2016}. If the mapping functions are cycle-consistent, for each image $x$ from domain $A$, the mapping cycle should bring $x$ back to a specified stage, i.e. $x\rightarrow Cpr(x) \rightarrow G_{A\rightarrow C}(x) \rightarrow G_{C\rightarrow B}(G_{A\rightarrow C}(x)) \approx Cpr(x)$. This means, to recover $Cpr(x)$ from $G_{C\rightarrow B}(x)$, the information contained in $Cpr(x)$ such as structure/texture should be preserved by $G_{A\rightarrow C}$. The cycle consistency loss is formulated as:
\begin{equation}\label{eq:cycleloss}
\begin{aligned}
  \mathcal{L}_{cyc}(M_1,M_2)&=\underset{x\sim \mathbb{P}_A}{\mathbb{E}} [\| M_2(M_1(Cpr(x)))-Cpr(x) \|] \\
  &+\underset{y\sim \mathbb{P}_C}{\mathbb{E}} [\|M_1(M_2(y))-y\|].
\end{aligned}
\end{equation}

\subsubsection{Full Objective}
Our overall objective is:
\begin{equation}\label{eq:objective}
\begin{aligned}
    \mathcal{L}(G_{A\rightarrow C},G_{C\rightarrow B},D_C,D_B)&=\mathcal{L}_{GAN}(G_{A\rightarrow C},D_C,A,C)\\
    &+\mathcal{L}_{GAN}(G_{C\rightarrow B},D_B,C,B)\\
    &+\lambda \mathcal{L}_{cyc}(M_1,M_2).
\end{aligned}
\end{equation}

Earth-Mover distance from WGAN \cite{wgan} is used to formulate objectives for optimization. In WGAN, discriminators are constrained to be 1-Lipschitz functions and their losses are constructed using the Kantorovich-Rubinstein duality.%\cite{Villani2009Optimal}.

There are works shows that the original WGAN tends to having extreme-valued or extremely distributed weights because of the weight clipping scheme \cite{improvewgan2017}. To alleviate this phenomenon, we optimize the expectation using softmax cross-entropy. The removal of the last activation in discriminator prevents the weights from growing too large. We argue that together with softmax cross-entropy, activation removal, batch normalization and data augmentation averting large weights caused by over-fitting, the loss can be an efficient approximation to K-Lipschitz function.

The adversarial loss of discriminator that distinguishes real samples in domain $C$ from fake samples generated by mapping functions $G_{A\rightarrow C}$ is:
\begin{multline}\label{eq:loss_da}
  \mathcal{L}_{D}(C,G_{A\rightarrow C}) = \underset{y\sim \mathbb{P}_C}{\mathbb{E}}  CE\left(S(D_C(x)),\vec{1} \right) +\\
  \underset{x\sim \mathbb{P}_A}{\mathbb{E}} CE\left(S(D_C( G_{A\rightarrow C}(x))),\vec{0} \right) ,
\end{multline}
where $CE$ stands for cross-entropy function and $S$ is softmax function.

The adversarial loss of generator that takes input from domain $A$ and generate samples in domain $C$ is:
\begin{equation}\label{eq:adv_loss_ga}
  \mathcal{L}_G(A,C)= \underset{x\sim \mathbb{P}_A}{\mathbb{E}} CE\left(S(D_C(G_{A\rightarrow C}(x))),\vec{1} \right).
\end{equation}

We learn the discriminative models via performing the minimization:
\begin{equation}\label{eq:objective_d}
J^{D^*} = \underset{\theta_D}{\min}~ \mathcal{L}_{D}(C,G_{A\rightarrow C})+\mathcal{L}_{D}(B,G_{C\rightarrow B}).
\end{equation}

The generators are learned by the following optimization:
\begin{equation}\label{eq:objective_g}
J^{G^*}= \underset{\substack{\theta_G}}{\min}~
\mathcal{L}_G(A,C)+
\mathcal{L}_G(C,B)+
\lambda \mathcal{L}_{cyc}(M_1,M_2).
\end{equation}

%%%%%%%%%%%%%%%%%%%%%%%%%%%%%%%%%%%%%%%%%%%%%%%%%%%%%%%%%%%%%%%%%%%%%%%%%%%%%
%%%%%%%%%%%%%%%%%%%%%%%%%%%%%%%%%%%%%%%%%%%%%%%%%%%%%%%%%%%%%%%%%%%%%%%%%%%%%
%%%%%%%%%%%%%%%%%%%%%%%%%%%%%%%%%%%%%%%%%%%%%%%%%%%%%%%%%%%%%%%%%%%%%%%%%%%%%

\subsection{Architecture}
\label{sec:arch}
\begin{figure}[!t]
    \centering
    \subfloat[]{
    \includegraphics[height=0.5in]{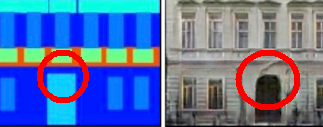}
    \label{fig:pix2pix_fronts}
    }
    \subfloat[]{
    \includegraphics[height=0.5in]{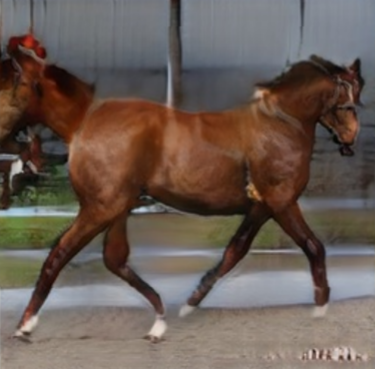}
    \label{fig:pgan_horse}
    }
    \subfloat[]{
    \includegraphics[height=0.5in]{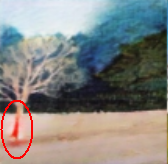}
    \label{fig:cyclegan_trees}
    }
    \subfloat[]{
    \includegraphics[height=0.5in]{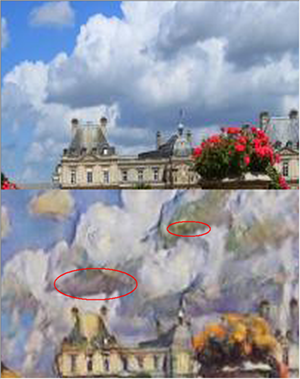}
    \label{fig:cyclegan_cloud}
    }
%    \subfloat[]{
%    \includegraphics[height=0.5in]{imgs/inconsistent_gan/synthesis_gan_wall.png}
%    \label{fig:sgan_wall}
%    }
%    \subfloat[lost contour]{
%    \includegraphics[height=0.7in]{imgs/inconsistent_gan/synthesis_gan_ground.png}
%    \label{fig:sgan_roads}
%    }
    \caption{Common problems in GANs.
    (a) Inconsistent structure in pix2pix. (b) Fabricated patterns in Progressive-GAN.
    (c) Incorrect coloring in Cycle-GAN. (d) Inconsistent coloring in Cycle-GAN.}
    \label{fig:inconsistent_gans}

\end{figure}

Prevailing architectures of generators like in most of cGANs \cite{cgan} are based on encoder-decoder structure for color stylization and transfer \cite{pix2pix,cyclegan}.
%This is partially because artistic stylization tasks require relatively strong re-creation based on the original input.
However, the auto-encoder architecture is prone to generating fabricated patterns/textures with inconsistent shapes and inconsistent coloring, as shown in Fig. \ref{fig:inconsistent_gans}, making it not suitable for visualization tasks which require precision and preciseness.
Some attempts have been made to use U-net to preserve the image structure \cite{pix2pix,dualgan} but the improvement is not significant.

The architecture of VGAN differs from previous GANs in two primary aspects. First, the images is not down-sampled or up-sampled spatially. This ensures that the original details are not destroyed by down-sampling and no extra texture is created by up-sampling. Second, the kernel size of all the convolutions are set to 1 to prevent the network from generating blurred texture.

\begin{figure}[!t]
\centering
\includegraphics[width=3.3in]{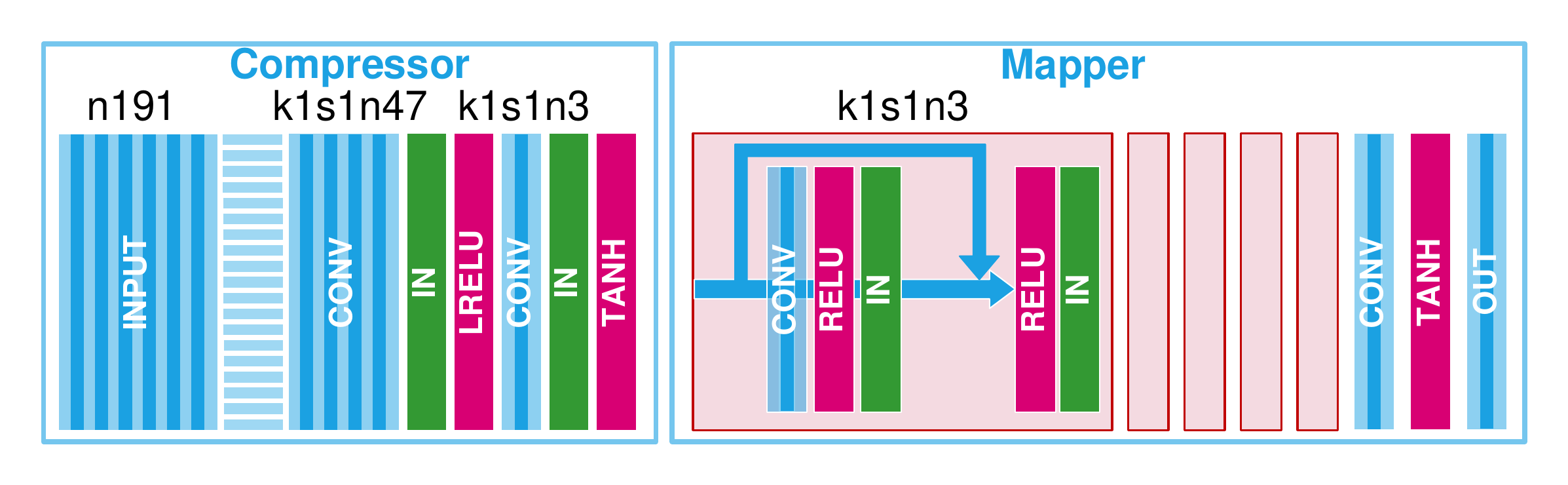}
\caption{Network architecture of the generator(ours).}
\label{fig:gen_ours}
\end{figure}

The architecture of generator $G_{A\rightarrow C}$ is shown in Fig. \ref{fig:gen_ours}. The network architecture of the mapper in generator $G_{C\rightarrow B}$ is identical to the mapper in $G_{A\rightarrow C}$.
The compressor network contains two stride-1 convolutions. The output of compressor is activated by $tanh$ so that the output is in the same range with the output of generator $G_{C\rightarrow B}$. The mapper network in generator $G_{A\rightarrow C}$ contains 5 residual blocks and one convolution layer with stride set to 1. We leverage the residual blocks in our architecture based on \cite{resnet2015}. Instance normalization is used in generator networks to diminish the influence of instance-specific contrast information in each input image \cite{2016instancenorm}.

\begin{figure}[!t]
\centering
\includegraphics[width=2.5in]{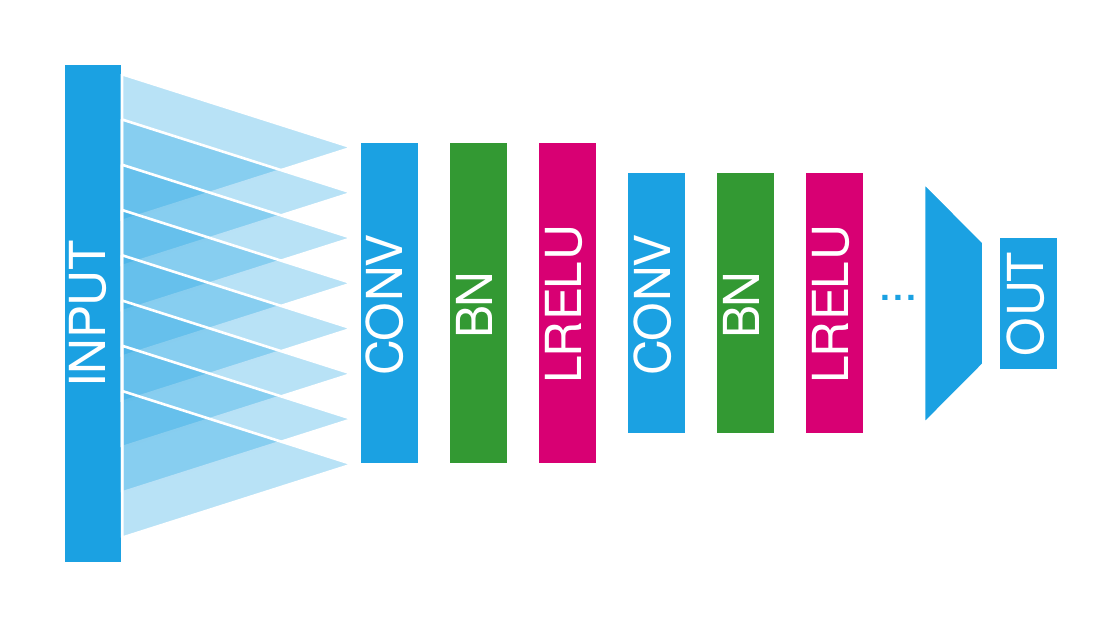}
\caption{Architecture of the discriminator.}
\label{fig:disc}
\end{figure}

The architecture of the discriminator is shown in Fig. \ref{fig:disc}. It consists of 4 convolutional layers with batch-normalization and Leaky-ReLU activation. Each convolutional layer reduce both the height and width of an image by a factor of 2. At the end of the network, the values given by the last convolution operation are reshaped to a vector as the output. For better gradient behaviour, the final output is not activated by any non-linear operation~\cite{wgan}.

%The two-generator structure is essential to our visualization architecture because the compressed 3-band image should contain color topology information and precise spatial structure so that the mapper could focus on (and is also designed to) transferring only the color instead of fabricating pixel topology or spatial texture.
%The topology is kept by imposing an explicit constraint on the network which requires the cycle consistent be minimized.

%Another implicit constraint is imposed by the pixel-wise architecture itself.
In our model, all the kernel size is set to 1. It is highly unlikely to generate extra fabricated artifacts and textures by such an pixel-wise computation.
On the contrary, pixel-wise operations are inclined to cause topology simplification and feature merging.
With the explicit cycle-consistency constraint reinforcing the information preservation and the implicit constraint reducing unwanted excessive creations, the networks is optimized to be structure consistent and nature in generation style.

We also tried a GAN model without the cycle architecture. The spatial structure in the images generated by the model were not well preserved. The colors in those visualization results were incorrect, either. In order to improve the performance, we explored several losses but did not observe significant improvement.

\section{Experiments}
\label{sec:experiments}

\subsection{Datasets}
\begin{figure}[t]
\centering
\includegraphics[angle=90,width=0.7\linewidth]{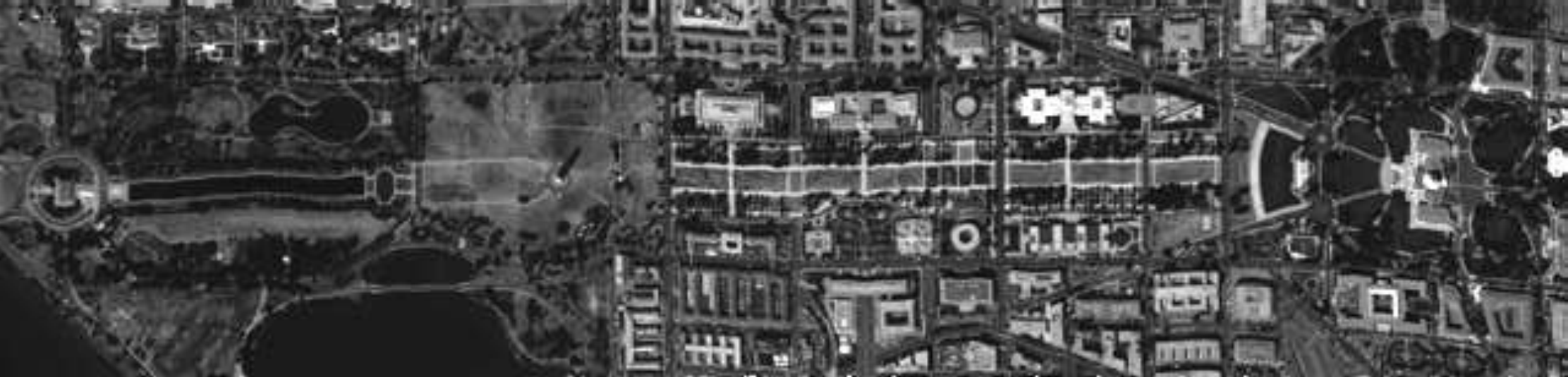}
\caption{The 50th band of the HSI}\label{fig:band50}
\end{figure}

\begin{figure}[t]
\centering
\includegraphics[width=0.7\linewidth]{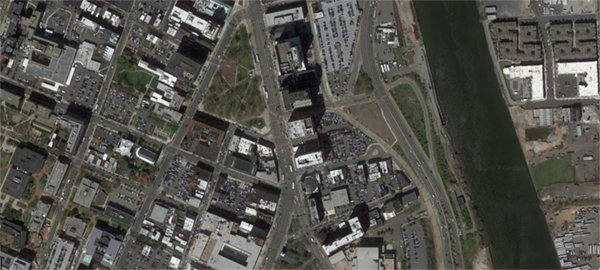}
\caption{One of the RGB images taken from Google Maps.(New York city)}
\label{fig:googlemap_newyork}
\end{figure}

In our experiment, we used a remote sensing hyperspectral image and a set of natural RGB images obtained  from Google Earth for training.
The hyperspectral image is taken over Washington D.C. mall by the Hyperspectral
Digital Imagery Collection Experiment sensor. The
data consist of $191$ bands after noisy bands are removed, in
which the size of each band image is $1208 \times 307$. Fig. \ref{fig:band50}  shows its 50th band image.
We collected 12 RGB images including scenes of New York(see Fig. \ref{fig:googlemap_newyork}), Orlando, and Washington. These images are of roughly the same size (about $1600 \times 800$) and have different spatial resolution.
The spectral images and RGB images are sliced into 6000 overlapped patches with the size of $128\times 128$ for training.
We augmented the data set by applying flipping and rotation on the patches.
For testing, the spectral image of D.C. mall is cut into non-overlapping patches.
The visualization result is obtained by stitching the non-overlapping patches together.

\subsection{Implementation Details}

We used ADAM for optimization and set the learning rate of the generator and the discriminator to 0.0001 and 0.00001, respectively.
The weight of cycle-loss $\lambda$ is set to 50.
In order for the generator to produce more diverse outputs, we follow the mini-batch technique proposed in ~\cite{traingan2016}, i.e., the discriminators are equipped with cached queues that store the most recent 50 fake images and 50 real images.
The epoch size is set to  6000 and the model is trained for 2 epochs using $A$ and $B$ training data with a batch size of 1.

\subsection{Visual and Quantitative Comparisons}

\begin{figure}[!t]
\centering
\subfloat[LP band selection]{
\includegraphics[angle=90,width=0.7\linewidth]{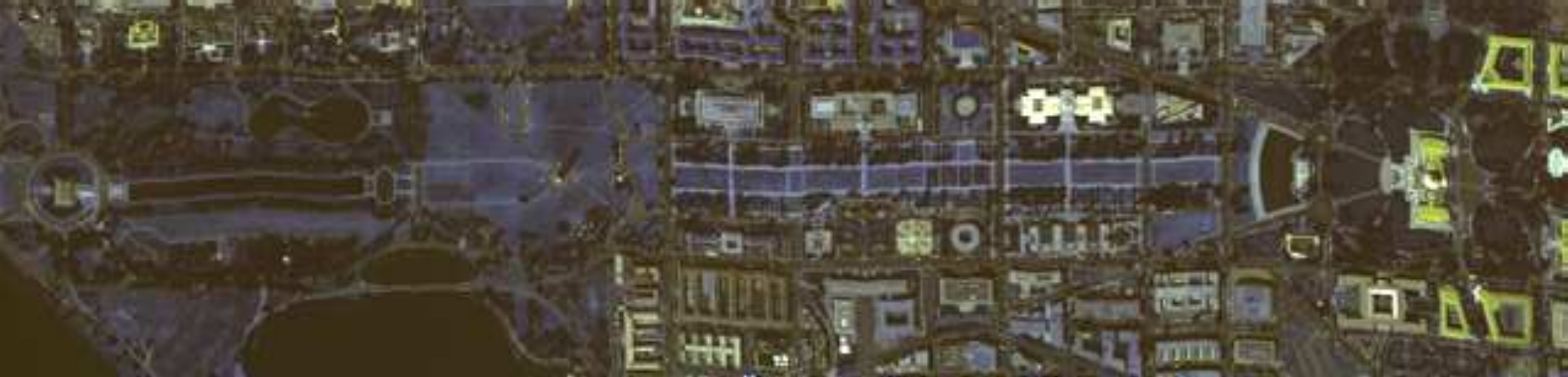}
\label{fig:DClpbandselection}
}
\\
\subfloat[Manifold alignment]{
\includegraphics[angle=90,width=0.7\linewidth]{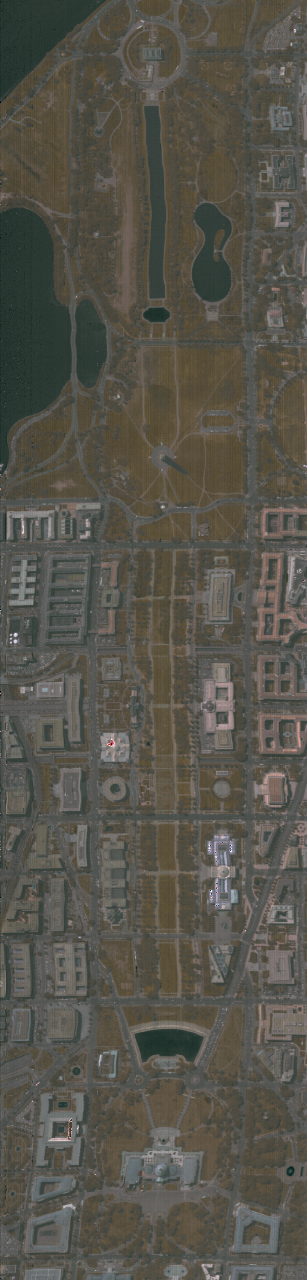}
\label{fig:DCmanifoldalignment}
}
\\
\subfloat[Stretched CMF]{
\includegraphics[angle=90,width=0.7\linewidth]{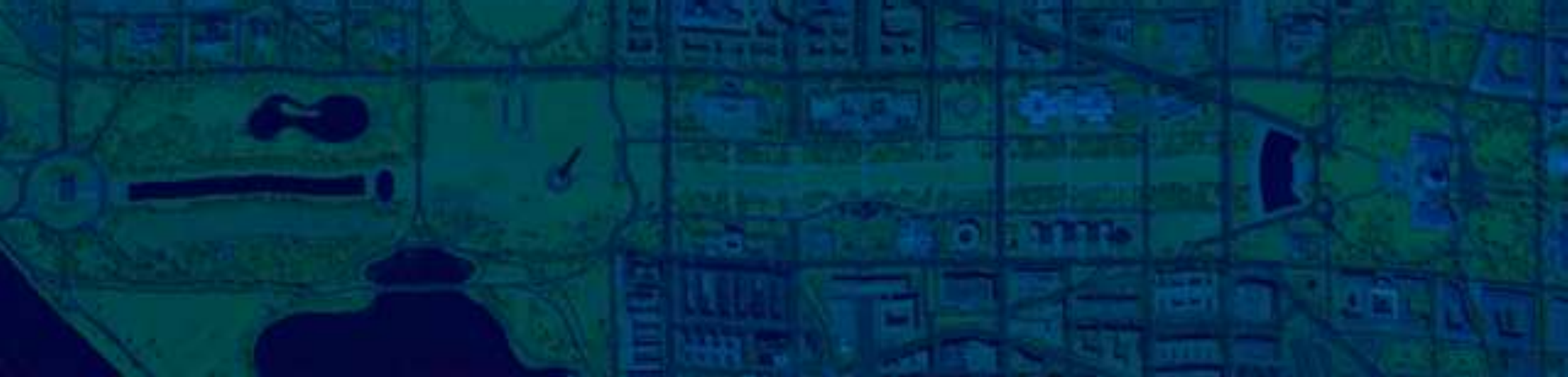}
\label{fig:DCCMF}
}
\\
\subfloat[Bilateral filtering]{
\includegraphics[angle=90,width=0.7\linewidth]{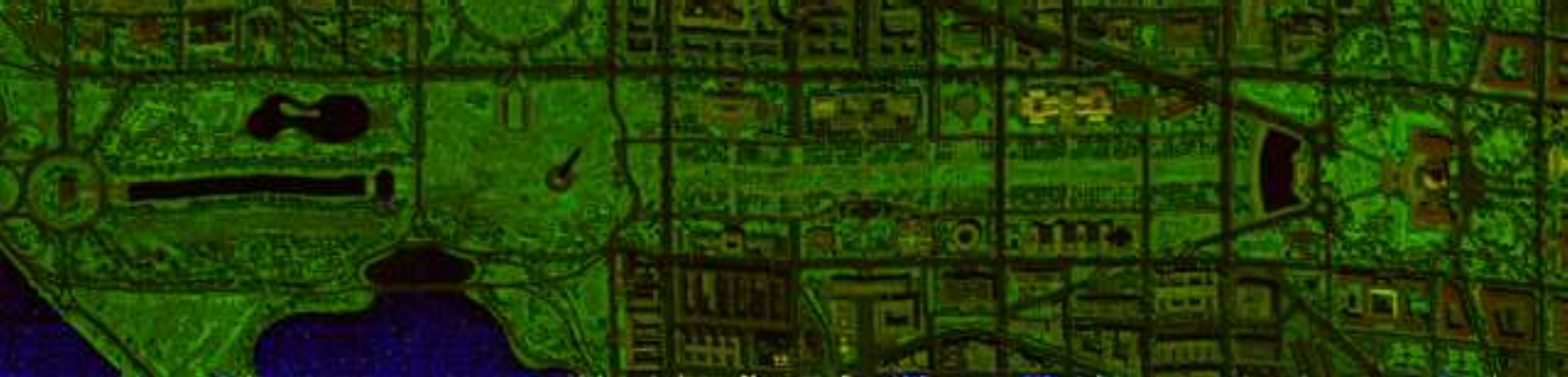}
\label{fig:DCBF}
}
\\
\subfloat[Bicriteria optimization]{
\includegraphics[angle=90,width=0.7\linewidth]{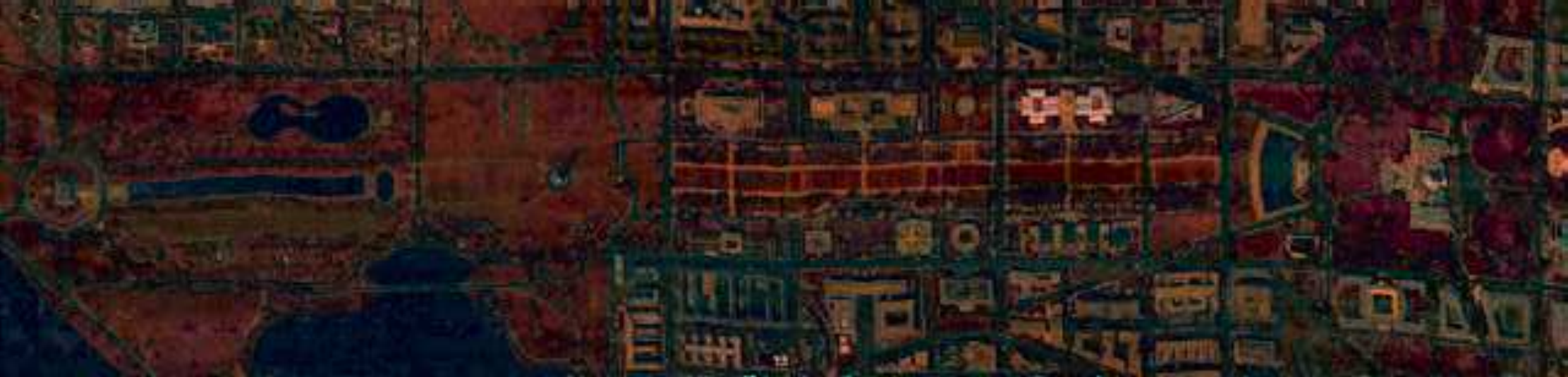}
\label{fig:DCBCOCDM}
}
\\
%\subfloat[Ours]{
%\includegraphics[angle=90,width=0.7\linewidth]{img/ours_C2A_pixnet.png}
%\label{fig:DCMLS}
%}
%\\
\subfloat[VGAN]{
\includegraphics[angle=180,width=0.7\linewidth]{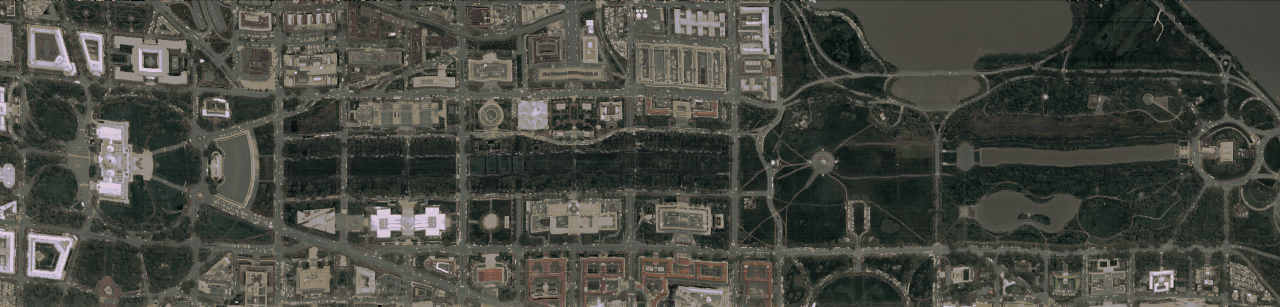}
\label{fig:GAN}
}
\\
\subfloat[Ground-truth RGB image]
{\includegraphics[angle=0,width=0.7\linewidth]{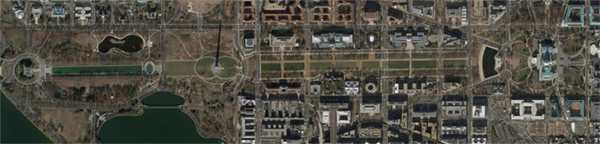}
\label{fig:registered2}
}
\caption{Visual comparison of different visualization approaches on the Washington DC Mall data set.
The spectral image of D.C. Mall was taken in Aug, 1995, and the RGB image from Google Earth was taken in 2017 winter.}
\label{fig:comparison}
\end{figure}

We compare our method with several methods including LP Band Selection \cite{su2014LP}, Stretched CMF \cite{cmf2005}, Bilateral Filtering \cite{kotwal2010visualization}, Bicriteria Optimization \cite{bicriteria2012} and Manifold Alignment \cite{liao2016hscolor}.
Fig. \ref{fig:comparison} shows the visual comparison of different visualization approaches on the Washington D.C. mall data.
It can be seen that the result of VGAN not only has a very natural tone but also preserves fine details.

Since there is no universally accepted standard for the quantitative assessment of spectral image visualization,  we adopt 4 metrics including entropy \cite{kotwal2010visualization}, root-mean-square error(RMSE) \cite{zhu2007evaluation}, separability of features \cite{cui2009interactive} and correlation coefficients between the RGB components (CORR) \cite{zhu2007evaluation}.

\emph{Entropy}: An image
with higher entropy contains richer information than ones with
low entropy. The entropy of a single-channel image is given by:
\begin{equation}\label{eq:entropy}
  h=-\sum_{x}p(x)\ln p(x),
\end{equation}
where $p(x)$ is the probability density of the intensity level $x$ in
the image. For an RGB image, its entropy is the average entropy
of all channels.

\emph{RMSE}: RMSE between the true color image and the visualized
image is a straight-forward way to evaluate whether the visualization
has natural colors~\cite{zhu2007evaluation}. The ground truth true color image is set to the RGB satellite image captured from the same location as the spectral image (Fig.~\ref{fig:registered2}).

Let $s(x)$ and $s'(x)$  denote the RGB vector of pixel $x$
in the visualized image and the true color image, respectively. The RMSE
between $s$ and $s'$ over the whole image is defined by:
\begin{equation}\label{eq:rmse}
  r = \sqrt{\frac{1}{N}\sum_x \left( s(x)-s'(x) \right)}.
\end{equation}

\emph{CORR}:
Since natural images have a high degree of correlation
between the RGB components~\cite{zhu2007evaluation}, the correlation between the RGB components of the
visualizations indicates the similarity of the visualizations to natural images.
The two-dimensional normalized correlation function for images is defined as:
\begin{equation}\label{eq:corr}
  CORR_{X,Y}=\frac{\sum_{m,n} (X_{mn}-\bar{X})(Y_{mn}-\bar{Y})}{\sum_{m,n}(X_{mn}-\bar{X})^2 \sum_{m,n} (Y_{mn}-\bar{Y})^2},
\end{equation}
where $X$, $Y$ are two image channel vectors; and $\bar{X}$, $\bar{Y}$ are the means
of $X$ and $Y$, respectively. $C_{XY}$ is a real number between -1
and 1.
the overall CORR is obtained as the mean value of $C_{R,G}$, $C_{R,B}$ and $C_{B,G}$.

\emph{Separability of Features}: Separability of features measures how
well distinct pixels are mapped to distinguishable colors. The
basic idea is that the average distance between two pixels in the
color space should be as large as possible \cite{cui2009interactive}.
Separability of features is defined as:
\begin{equation}\label{eq:separability}
  \delta= \frac{1}{(N-1)^2}\sum_{x\neq y}d(x,y),
\end{equation}
in which, $d(x,y)$ is the Euclidean distance between the pair of
pixels $x$ and $y$ in the RGB space and $N$ is the number of pixels.
$\delta$ denotes the average pairwise Euclidean distance in terms of
all pixel pairs. Larger $\delta$ yields better separability of
features.

The quantitative comparison results are shown in Table. \ref{table:performance}.
Comparisons on entropy and RMSE shows that manifold alignment method's performance stands out by solving the tradeoff between information preservation and natural rendering.
Without the requirement of paired labeling, our method still performs well under entropy and RMSE metrics. The results indicate that VGAN can visualize images with rich information and similar color to the unseen ground-truth.
Compared to manifold alignment, which requires a set of matching pixels for color transfer, our method achieves better CORR, which indicates the strong capability of VGAN in producing visualizations with natural color distribution. Also, the separability of our approach ranks the first place.
This means, our approach can render the image in natural color and, at the same time, preserve the pixel separability in a way that doesn't need extra explicit constraints or supervision.

Notably, form the results shown in Fig. \ref{fig:GAN} we can see that even when the real-color training samples and the spectral image were taken in different seasons, our method is able to learn the most likely appearance of trees in summer instead of the sallow color directly transferred by manifold alignment.
In our experiments we found that, the network without cycle-loss performs much worse in terms of separability. This meets our expectation since cycle-loss requires the separability be preserved so that the details can be best recovered.

\begin{table}[t]
\centering
 \caption{Performance comparisons.}
  \label{table:performance}
  \begin{tabular}{ccccccc}
    \hline
    Method                   & Entropy        & RMSE           & CORR          & Separability           \\
    \hline
    LP Band Selection        & 6.59           & 68.35          & 0.92          & 55.68                  \\
    \hline
    Stretched CMF            & 3.89           & 92.46          & 0.36          & 30.08                  \\
    \hline
    Bilateral Filtering      & 3.31           & 102.43         & 0.73          & 47.54                  \\
    \hline
    Bicriteria Optimization  & 5.75           & 98.47          & 0.63          & 52.43                  \\
    \hline
    Manifold Alignment       & \textbf{10.32} & \textbf{33.23} & 0.89          & 34.64                  \\
    \hline
    Ours                     & 8.25           & 49.99          & \textbf{0.98} & \textbf{64.89}         \\
    \hline
  \end{tabular}
\end{table}

\section{Conclusion}
\label{sec:conclusion}
In this paper we presented an unsupervised end-to-end VGAN for displaying spectral images in natural colors.
By minimizing the adversarial loss of the discriminator that is trained to differentiate between false-color images and natural-color images, VGAN is able to generate natural-looking visualizations.
In addition, we use a cycle-loss to overcome the ambiguity and structure inconsistent problem of classic GAN.
Visualization results show that VGAN is capable of producing visualizations with rich information and very natural color distributions.
Pixels in the spectral images are also well separated in the RGB space.


%\section*{Future Works}



\bibliographystyle{IEEEtran}
\bibliography{mybib}

\end{document}